# How Trustworthy Are LLM-as-Judge Ratings for Interpretive Responses? Implications for Qualitative Research Workflows

Songhee Han[0000-0002-8984-7560], Jueun Shin[0009-0007-6830-9791], Jiyoon Han[0009-0002-6389-0445], Bung-Woo Jun[0009-0009-7336-2532], and Hilal Ayan Karabatman[0000-0002-8925-7341]

Florida State University, Tallahassee, FL 32306, USA
songhee.han@fsu.edu

**Abstract.** As qualitative researchers show growing interest in using automated tools to support interpretive analysis, a large language model (LLM) is often introduced into an analytic workflow *as is*, without systematic evaluation of interpretive quality or comparison across models. This practice leaves model selection largely unexamined despite its potential influence on interpretive outcomes. To address this gap, this study examines whether LLM-as-judge evaluations meaningfully align with human judgments of interpretive quality and can inform model-level decision making. Using 712 conversational excerpts from semi-structured interviews with K–12 mathematics teachers, we generated one-sentence interpretive responses using five widely adopted inference models: Command R+ (Cohere), Gemini 2.5 Pro (Google), GPT-5.1 (OpenAI), Llama 4 Scout-17B Instruct (Meta), and Qwen 3-32B Dense (Alibaba). Automated evaluations were conducted using AWS Bedrock's LLM-as-judge framework across five metrics, and a stratified subset of responses was independently rated by trained human evaluators on interpretive accuracy, nuance preservation, and interpretive coherence. Results show that LLM-as-judge scores capture broad directional trends in human evaluations at the model level but diverge substantially in score magnitude. Among automated metrics, Coherence showed the strongest alignment with aggregated human ratings, whereas Faithfulness and Correctness revealed systematic misalignment at the excerpt level, particularly for non-literal and nuanced interpretations. Safety-related metrics were largely irrelevant to interpretive quality. These findings suggest that LLM-as-judge methods are better suited for screening or eliminating underperforming models than for replacing human judgment, offering practical guidance for systematic comparison and selection of LLMs in qualitative research workflows.

## 1. Introduction

An increasing number of qualitative research workflows adopt large language models (LLMs), where they are used to summarize interviews, generate interpretive codes, and assist with thematic analysis. These developments have accelerated interest in whether LLMs can reliably perform interpretive tasks traditionally handled by trained human coders. Recent advances in LLM-as-judge methods, in which one LLM evaluates the output of another, have shown promising reliability in highly structured domains such as clinical summarization. For example, Croxford et al. (2025) reported that a medical LLM-as-judge achieved human-level agreement when evaluating summaries of electronic health records. The authors argued that LLM judgments could approximate human ratings when they were carefully designed and implemented. Similarly, Anghel et al. (2025b) noted the growing adoption of LLM-as-judge uses across tasks; however, they highlighted that issues of bias, prompt sensitivity, and instability presented unresolved gaps for domains that demand nuanced interpretive reasoning.

Interpretive analysis requires sensitivity to nuance, context, and speaker intent (Austin & Sutton, 2014). Particularly, interview-based interpretation goes beyond surface-level correctness; researchers must infer meaning beyond literal wording, account for fragmented or incomplete utterances, and distinguish how both interviewers and interviewees shape conversational trajectories. These demands differ substantially from the more structured tasks, such as factual summarization or correctness scoring, typically used to benchmark the reliability of LLMs as judges (Islam et al., 2024; Luo et al., 2024). Existing empirical studies also emphasize that LLM-as-judge performance is highly task-dependent; fine-tuned judges often overfit to narrow evaluation schemes and generalize poorly outside their training domain (Jiang et al., 2025).

Altogether, these findings indicate that while LLM-as-judge methods have demonstrated promise in constrained technical settings, it remains unclear whether they can meaningfully support model-level decision making and interpretive evaluation in qualitative research contexts. This gap is particularly notable given that qualitative research practitioners increasingly rely on a single inference model for interpretive work, often without systematic comparison or clear guidance on how automated evaluation signals should be used.

To address this gap, we investigated whether AWS Bedrock's LLM-as-judge scoring could reliably evaluate the interpretive responses generated by multiple inference models. In this study, we focused on one-sentence interpretive responses, concise statements that captured what speakers were trying to communicate within the full conversational context. To decide which inference models to include, we considered whether each model was open-source or proprietary and its current level of use, prioritizing models commonly adopted by practitioners. Therefore, the evaluation set included five inference models available as of November 2025: Command R+ (Cohere), Gemini 2.5 Pro (Google), GPT-5.1 (OpenAI), Llama 4 Scout-17B Instruct (Meta), and Qwen 3-32B Dense (Alibaba). All models were tested using default or default-aligned decoding parameters to reflect realistic practitioner use. Through this design, we sought to assess the extent to which automated scores align with human judgments, identify which evaluation metrics are most informative for inference-model selection, and examine cases where human and automated evaluations diverge, with the goal of providing actionable guidance for qualitative research practitioners.

This study was guided by the three research questions (RQs):

1. To what extent do LLM-as-judge evaluations of interpretive responses align with human judgments of interpretive quality?

2. Which LLM-as-judge evaluation metrics show the strongest alignment with aggregated human ratings at the inference-model level, informing model selection for qualitative analysis?

3. What interpretive patterns emerge in excerpts where human and LLM-judge evaluations differ the most?

**2. Related Work**

2.1 LLM Use in Qualitative Research

Across different research disciplines, LLMs are increasingly applied to qualitative tasks such as summarizing texts, generating first-cycle codes, supporting grounded-theory analysis, and interpreting patterns in learner or teacher dialogue (Barany et al., 2024; Croxford et al., 2025; Liu & Sun, 2025; Shin, 2025; Yue et al., 2025). Emerging work suggests that LLMs can produce coherent interpretive statements, but concerns persist regarding faithfulness, hallucination, and sensitivity to contextual nuance (Chen et al., 2024; Khan et al., 2024), issues that are especially important in interview-driven research traditions. Because interpretive accuracy is tied to meaning preservation rather than surface-level content similarity, automated evaluation requires more than standard natural language processing metrics (Chen et al., 2024; Dai et al., 2023; Roberts et al., 2024).

Despite the growing interest in LLM-assisted qualitative analysis, many studies adopt a convenience-driven approach in which a single, widely accessible model, often GPT-based, is applied directly under default settings or with minimal prompt engineering. In most cases, researchers use an LLM "as is," without formally evaluating how well the model handles the specific linguistic, contextual, or speaker-related characteristics of their qualitative dataset. Only a small number of recent studies have begun to interrogate model performance more systematically (e.g., Chen et al., 2024; Khan et al., 2024; Wen et al., 2025). For example, Wen et al. (2025) compared the thematic-analysis performance of GPT-4o and GPT-4o-mini and found substantive differences in stability and agreement across models, demonstrating that model choice significantly affects interpretive quality. Their work illustrated the kind of testing that has been largely absent from much of the current literature, where interpretive reliability is frequently assumed rather than empirically established. This gap highlights the need for more systematic evaluation of LLM behavior before adopting models as qualitative coding assistants.

2.2 LLM-as-Judge Evaluation

LLM-as-judge approaches have been commonly framed as a promising solution for evaluating open-ended outputs using rubrics, particularly in settings where human evaluation is costly or difficult to scale (Li et al., 2025). Empirical studies in educational and assessment domains report meaningful alignment between LLM-based scores and human judgments when tasks are structured and evaluation criteria are explicit, such as short-answer scoring, reading comprehension, and essay assessment (Atkinson & Palma, 2025; Clauser et al., 2024; Henkel et al., 2025). To facilitate evaluation work, some studies have introduced multidimensional and rubric-guided evaluation frameworks that emphasize transparency, consistency, and reproducibility in automated judging. For example, Anghel et al. (2025a) propose a graph-based multi-LLM framework designed to make scoring processes more interpretable and reproducible across evaluators. Therefore, these studies collectively suggest that LLM-as-judge systems can approximate human evaluation under constrained conditions where assessment criteria are well defined.

However, prior research has also identified important limitations of LLM-based automated assessment. A recent systematic review synthesizing 49 studies reported that although models such as GPT-4 often achieved high agreement with human evaluators, issues related to algorithmic bias, hallucinations, and transparency remained (Emirtekin, 2025). Additionally, Jiang et al. (2025) raised concerns about reliability and robustness issues by reporting that LLM-as-a-Judge for coding remained limited by significant randomness, order sensitivity, and inconsistent judgments across models and tasks. These findings suggest that while LLM-as-judge approaches show promise in structured assessment contexts, their reliability and generalizability beyond such contexts remain insufficiently examined.

3. Method

**3.1. Dataset and Preprocessing**

The dataset consisted of eight semi-structured interviews with K–12 mathematics teachers. The interviews focused on teachers' conceptions of critical thinking, their views on the instructional affordances of AI, and the professional learning needs associated with AI integration. Transcripts were systematically preprocessed by two qualitative researchers to remove speech fillers and non-meaning-bearing segments while preserving complete conversational turns. Both the interviewer's and interviewee's talks were retained to support an accurate interpretation of meaning within the broader discourse context. After preprocessing, the full dataset consisted of 712 meaning-bearing conversational excerpts.

### 3.2. LLM Interpretive Output Generation

All inference models generated interpretive outputs for the entire preprocessed transcript. Each model received the full conversation to ensure consistent access to contextual information. For every excerpt, the models were prompted with the following instruction:

> Task: For the EXCERPT below, generate an interpretive response in exactly one sentence. The interpretive response should convey what the speaker is trying to express, not repeat or summarize the exact wording. Output only the single sentence with no explanation or preamble.

Bedrock-hosted models (Command R+, Llama 4 Scout-17B Instruct, and Qwen 3-32B Dense) were run using the decoding parameters that Bedrock applies by default. These defaults reflect the platform's operational behavior and represent the conditions most encountered by practitioners who rely on out-of-the-box model usage.

Two models were not available through Bedrock; therefore, they were accessed through their native APIs: Gemini 2.5 Pro and GPT-5.1. Because the model providers do not publish fixed decoding defaults, we adopted researcher-aligned settings that follow a stable pattern widely used in LLM research: *temperature* = 0.5, *top-p* = 0.9, and *max_tokens* = 512. This alignment was guided by the Meta-Llama defaults published in AWS Bedrock's model documentation ([AWS, n.d.](#)). Although these defaults are specific to Meta-Llama models, they offer a reasonable benchmark for aligning GPT-5.1 and Gemini 2.5 Pro in the absence of provider-disclosed decoding defaults, therefore ensuring cross-model comparability.

### 3.3 LLM-as-Judge Evaluation Procedure

Automated evaluation was conducted using Claude 3.5 Sonnet through AWS Bedrock's LLM-as-judge service. In Bedrock, two evaluation modes are available: an auto-evaluation for Bedrock-hosted models and a bring-your-own (BYO) evaluation for models accessed externally (AWS, n.d.). In this study, three models (Command R+, Llama 4 Scout-17B Instruct, Qwen 3-32B Dense) were evaluated through Bedrock's auto-evaluation pipeline, whereas Gemini 2.5 Pro and GPT-5.1 were evaluated through BYO mode. Claude 3.5 Sonnet was chosen as the LLM judge because prior research has demonstrated that it has strong alignment with human judgments in rubric-based evaluation settings (Cipriano et al., 2025).

For the purposes of this study, five evaluation metrics were selected by us: Faithfulness, Correctness, Coherence, Harmfulness, and Stereotyping. Table 1 indicates each metric's definition (AWS, n.d.). These selections were based on their relevance to interpretive-response quality. These metrics were then used by the LLM judge to automatically score all model-generated responses. In addition to numerical scores, the LLM judge also returned explanation text associated with each metric score, providing brief descriptions of how the scoring rubric was applied to the inference model-generated interpretation. In total, the LLM judge produced metric scores for 3,560 interpretive responses (712 excerpts × 5 inference models). The resulting evaluation scores were used to examine the extent to which LLM-as-judge scoring aligns with human judgments on interpretive quality. Explanation text associated with each score was retained to support subsequent qualitative examination of the LLM judge's scoring. Table 2 shows examples of interpretive responses from LLMs for the same original excerpts.

**Table 1. AWS Bedrock LLM-as-Judge Evaluation Metrics and Definitions**

| Metric | AWS Definition |
|---|---|
| Faithfulness | Assesses whether the model's response remains grounded in the provided context, avoiding unsupported or hallucinated information. |
| Correctness | Evaluates whether the response is factually and logically correct given the task requirements or reference information. |
| Coherence | Measures the clarity and logical consistency of the response, including whether ideas are organized and free of contradictions. |
| Harmfulness | Identifies the presence of harmful, unsafe, or inappropriate content (e.g., violence, hate speech, abusive language). |
| Stereotyping | Detects biased or generalized statements about individuals or groups based on stereotypes. |

**Table 2. Examples of LLM-Generated Interpretive Responses for Two Consecutive Interview Excerpts.** * Gemini 2.5 Pro frequently produced truncated, mid-sentence outputs, even for short prompts and high token limits. This is one of many examples.

| Model | Interpretive response for Excerpt A (Interviewer: *What is mathematical critical thinking? I want to encourage you to think about it in your classroom situation. What kind of actions or what kind of talk from students signal critical thinking in math happens?*) | Interpretive Response for Excerpt B (Interviewee5: *So, I hear you, you're asking me, like, my definition of it, also what are some student actions that I see when they're participating? So... I think I would define math critical thinking as the ability to approach a problem, apply, or even taking something that you've already learned and, applying it to a new situation, but also being able to approach a problem.*) |
|---|---|---|
| Command R+ | Interviewer seeks to understand the respondent's perspective on mathematical critical thinking, prompting them to reflect on their classroom experiences and identify | Approaching a problem with a flexible mindset, drawing on prior knowledge, and adapting it to novel situations is what I understand to be |

| | student behaviors or discussions that indicate critical thinking in mathematics. | the essence of math critical thinking. |
|---|---|---|
| Gemini 2.5 Pro | Please describe the specific, observable student behaviors that you recognize as evidence of mathematical* | The speaker defines math critical thinking as the ability to transfer existing knowledge to solve unfamiliar problems. |
| GPT-5.1 | The interviewer is asking the teacher to reflect on real classroom experiences to identify specific student behaviors and comments that show they are engaging in genuine critical thinking about mathematics. | They're trying to describe math critical thinking as using what you've already learned to tackle new problems by thoughtfully figuring out how to approach and apply ideas in unfamiliar situations. |
| Llama 4 Scout-17B | The speaker is trying to understand how to identify and foster critical thinking in mathematics, specifically by reflecting on classroom experiences and student behaviors that indicate critical thinking is taking place. | The speaker is trying to clarify that they understand the question as asking for both a definition of math critical thinking and examples of student behaviors that demonstrate it. |
| Qwen 3-32B | The speaker is asking how to recognize mathematical critical thinking through student actions and dialogue in a classroom setting. | The speaker is trying to express that math critical thinking involves applying prior knowledge to new problems and approaching them with analytical skill. |

### 3.4. Human Rating Procedure

Human evaluation was conducted on a stratified subset of inference-model-generated interpretive responses. A total of 60 model-generated interpretations were selected using a stratified purposive sampling strategy for human evaluation (Palinkas et al., 2015). Sampling considered four linguistic and conversational variables: speaker role (interviewer vs. interviewee), utterance length (short, medium, long), sentence completeness (complete vs. fragmented), and interviewee identity across all eight teachers. Because each sampled excerpt was paired with one LLM-generated interpretation, this approach also captured variation attributable to model identity without requiring all five model outputs for each excerpt. This maximum-variation, stratified sampling approach is consistent with established qualitative research standards (Creswell, 2013; Patton, 2015) and with recent AIED work that evaluates human–LLM alignment using sampled, balanced subsets rather than entire corpora (e.g., Simon et al., 2025). Table 3 shows an overview of the sampled excerpt characteristics.

Four trained human evaluators participated in the study. Each of the 60 sampled inference model outputs was evaluated by three of the four evaluators, yielding 180 total rating assignments and distributing the workload evenly so that each evaluator assessed 45 items. The evaluators were provided access to the full transcript surrounding each sampled excerpt to make sure that their judgments reflected the same contextual information available to the inference models.

Evaluators applied the following three criteria to score each model-generated interpretive output: interpretive accuracy, nuance preservation, and interpretive coherence (Table 4). Table 4 presents the full rubric provided to the human evaluators, summarizing the definitions, expectations, and scoring anchors for each criterion. These criteria reflect established guidance in qualitative research regarding credibility, depth of interpretation, and analytic clarity (Lincoln & Guba, 1985; Miles et al., 2014; Saldaña, 2021), ensuring that human evaluations provided a consistent and theoretically grounded benchmark for validating LLM-as-judge scoring.

**Table 3. Characteristics of the Sampled Excerpts ($n = 60$).** *Counts reflect the total number of interviewee excerpts, with representation from Interviewees 1–8 in the following distribution: 2, 6, 2, 6, 4, 3, 3, 4.

| Characteristic | Categories | Count | Percentage |
| --- | --- | --- | --- |
| Speaker role | Interviewee | 30* | 50% |
|  | Interviewer | 30 | 50% |
| Excerpt length (token-based) | Short (2–10 tokens) | 18 | 30% |
|  | Medium (11–40 tokens) | 22 | 36.7% |
|  | Long (41–364 tokens) | 20 | 33.3% |
| Sentence completeness | Complete | 42 | 70% |
|  | Fragment | 18 | 30% |
| Inference model | Command R+ | 12 | 20% |
|  | Gemini 2.5 Pro | 12 | 20% |
|  | GPT-5.1 | 12 | 20% |
|  | Llama 4 Scout-17B Instruct | 12 | 20% |
|  | Qwen 3-32B Dense | 12 | 20% |

**Table 4. Rubric for Human Evaluation of LLMs' Interpretive Outputs.** A score of 3 indicates a mid-level interpretation.

| Criterion | Definition | Indicators of low scores (1–2) | Indicators of high scores (4–5) |
|---|---|---|---|
| Interpretive accuracy (Lincoln & Guba, 1985 - credibility, faithful representation) | Extent to which the output captures the speaker's intended meaning. | Meaning is distorted, misrepresented, or only partially captured; interpretation contradicts or overlooks core intent. | Meaning is faithfully captured; response reflects the speaker's intended idea with conceptual accuracy. |
| Nuance preservation (Saldaña, 2021 - attention to layered meaning and interpretive depth) | Degree to which subtle, implicit, or secondary meaning cues are retained in the interpretation. | Response is generic, surface-level, or loses important nuance, context, or implicit meaning. | Response reflects layered meaning, implied intentions, or contextual subtleties beyond the literal text. |
| Interpretive coherence (Miles et al., 2014 - analytic clarity and coherence) | Clarity, structure, and appropriateness of the interpretation as a single, well-formed sentence. | Response is unclear, disorganized, confusing, or awkward; interpretation lacks internal consistency. | Response is concise, coherent, and contextually appropriate; reads as a clear interpretive statement. |

### 3.5. Analyses

Analyses proceeded in three stages. First, we assessed interrater reliability for the three human-rating criteria using the intraclass correlation coefficient (ICC; Shrout & Fleiss, 1979). Because the rotating triad design assigned different evaluators to each excerpt, we used ICC(3,k). Reliability of individual ratings was modest (ICC(3,1) = .492–.545), but the reliability of the averaged ratings was acceptable across criteria (Table 5). Therefore, each category was represented by the mean of the three evaluators' scores (Koo & Li, 2016).

To address RQ1, we examined alignment between LLM-as-judge scores and human judgments at the inference model level. Human ratings were analyzed both as separate three criteria (interpretive accuracy, nuance preservation, interpretive coherence) and as a composite interpretive-quality score obtained by averaging the three criteria. Spearman rank correlations were used to assess whether automated scores produced similar relative orderings of interpretive quality as human evaluators, including both a composite to composite and each metric levels. To assess the magnitude of differences, mean absolute error (MAE) was calculated with automated scores linearly rescaled to the human 1–5 rating scale.

To address RQ2, we compared aggregated human evaluations across the five inference models to assess model-level differences in interpretive output quality. We then examined metric-specific correlations between individual LLM-as-judge metric (Faithfulness, Correctness, Coherence, Harmfulness, Stereotyping) and human-rated three criteria to identify which automated metrics provided the strongest human-aligned signal.

Finally, to address RQ3, we identified excerpts where human and automated evaluations diverged most strongly by computing discrepancy scores between human evaluations and LLM-as-judge metrics. Excerpts with the largest discrepancies were selected for further review. For these high-discrepancy cases, we examined the original utterance, the inference model-generated interpretive responses, and the explanation text associated with the LLM judge's automated scores to determine the types of interpretive issues associated with human-LLM judge rating divergence.

**Table 5. Interclass Correlation Coefficient for Human Evaluations Criteria.**

| Criterion | ICC(3,1) | ICC(3,k) | Interpretation (Ave. ratings) |
|---|---|---|---|
| Interpretive accuracy | 0.545 | 0.782 | Good |
| Nuance preservation | 0.492 | 0.744 | Acceptable |
| Interpretive coherence | 0.492 | 0.744 | Acceptable |

## 4. Results

### 4.1 RQ1: Alignment Between LLM-as-Judge Evaluations and Human Judgments

Across models, the composite human score and composite automated scores showed moderate rank-order alignment (Spearman $\rho = .60$, $p = .28$) with the MAE of 0.91. Table 5 reports model-level composite human scores and composite LLM-as-judge scores for each inference model, providing context for the overall alignment analyses.

Across the three human rating criteria, correlations with LLM-as-judge metrics were generally modest and positive (Table 7), indicating limited but consistent directional correspondence. The strongest associations for interpretive accuracy and nuance preservation were observed with the LLM judge's Coherence metric ($\rho = .63$ and $\rho = .67$, respectively), although none of these reached statistical significance given the small number of models ($N = 5$). Relationships with Harmfulness and Stereotyping were negative across all human criteria. Overall, RQ1 results indicate that LLM judge's evaluations captured broad directional trends of human evaluations, but notable differences remained in score magnitude.

**Table 6. Human and LLM-as-Judge Scores Per Model**

| | Human evaluation criteria | | | | LLM-as-judge metrics | | | | | |
|---|---|---|---|---|---|---|---|---|---|---|
| Model | Interpretive accuracy | Nuance preservation | Interpretive coherence | Composite (rank) | Harmfulness | Faithfulness | Correctness | Coherence | Stereotyping | Composite* (rank) |
| commandRplus | 3.250 | 3.028 | 4.167 | 3.481 (5) | 1.011 | 4.251 | 4.371 | 4.997 | 1.011 | 4.540 (4) |
| gemini_2.5pro | 3.611 | 3.083 | 4.056 | 3.583 (4) | 1 | 3.728 | 3.815 | 4.882 | 1.022 | 4.141 (5) |
| gpt_5_1 | 3.833 | 3.528 | 4.778 | 4.046 (1) | 1 | 4.573 | 4.893 | 5 | 1 | 4.822 (3) |
| llama4_scout_17b | 3.250 | 3.222 | 4.861 | 3.778 (3) | 1.006 | 4.820 | 4.958 | 4.994 | 1.006 | 4.924 (2) |
| qwen | 3.750 | 3.250 | 4.833 | 3.944 (2) | 1.006 | 4.851 | 4.963 | 5 | 1.006 | 4.938 (1) |

*Note.* Human evaluations were on a 1–5 scale; LLM-as-judge ratings were linearly rescaled from 0–1 to 1–5 for comparability. All values are model-level averages. Composites of human and automated scores represent the mean of their respective domains. *This Composite includes only Faithfulness, Correctness, and Coherence because the others were negatively related to all human evaluation criteria.

### 4.2 RQ2: Usefulness of Individual LLM-as-Judge Metrics

As shown in Table 7, different automated metrics varied substantially in their correspondence to human evaluations. For interpretive accuracy and nuance preservation, the LLM judge's Coherence metric demonstrated the strongest alignment, with the highest correlations ($\rho = .63$, MAE = 1.44 and $\rho = .67$, MAE = 1.75). This pattern also extended to the human composite score, where Coherence again yielded the strongest correspondence ($\rho = .67$, MAE = 1.21). These results indicate that Coherence provides the most reliable automated estimate of overall interpretive response quality, although the MAE values indicate the differences remained non-negligible.

For interpretive coherence, the Faithfulness and Correctness metrics showed very strong correspondence (both $\rho = .90$, $p = .04$) alongside substantially smaller MAE values (0.14 and 0.16). By contrast, the Coherence metric showed weaker alignment with human coherence judgments ($\rho = .36$), despite a relatively small MAE (0.44).

Across all human metrics, Harmfulness and Stereotyping demonstrated negative or negligible correlations and consistently large MAE values, including for the composite, indicating that these metrics do not meaningfully reflect interpretive quality.

**Table 7. Alignment Between Human Evaluations and LLM-as-Judge Metrics**

| Human criterion | Automated metric | Spearman ρ | p-value | MAE |
|---|---|---|---|---|
| Interpretive accuracy | Faithfulness | 0.21 | .74 | 0.91 |
| | Correctness | 0.21 | .74 | 1.06 |
| | **Coherence** | **0.63** | .25 | 1.44 |
| | Harmfulness | –0.68 | .21 | 2.53 |
| | Stereotyping | –0.55 | .33 | 2.53 |
| Nuance preservation | Faithfulness | 0.60 | .28 | 1.22 |
| | Correctness | 0.60 | .28 | 1.38 |

|  | Coherence | **0.67** | **.22** | **1.75** |
|  | Harmfulness | –0.53 | .36 | 2.22 |
|  | Stereotyping | –0.87 | .05 | 2.21 |
| Interpretive coherence | **Faithfulness** | **0.90** | **.04** | **0.14** |
|  | **Correctness** | **0.90** | **.04** | **0.16** |
|  | Coherence | 0.36 | .55 | 0.44 |
|  | Harmfulness | 0.26 | .67 | 3.53 |
|  | Stereotyping | –0.67 | .22 | 3.53 |
| Composite | Faithfulness | 0.60 | .28 | 0.68 |
|  | Correctness | 0.60 | .28 | 0.83 |
|  | **Coherence** | **0.67** | **.22** | **1.21** |
|  | Harmfulness | –0.53 | .36 | 2.76 |
|  | Stereotyping | –0.87 | .05 | 2.76 |

**4.3 RQ3: Interpretive Issues in High-Divergence Excerpts**

Although LLM judge's Coherence metric showed most useful alignment with human ratings at the aggregated inference-model level (RQ2), excerpt-level analyses revealed that it provided little diagnostic value for identifying interpretive divergence: 59 of 60 received a score of 1.0. Explanation text accompanying these scores typically indicated that responses were considered "logically cohesive" whenever no contradictions or explicit reasoning errors were present. Because the metric functioned primarily as a binary indicator and did not differentiate among interpretations of varying quality, Coherence did not support discrepancy-based case selection. In contrast, Faithfulness and Correctness exhibited substantial variability and revealed systematic patterns in excerpts where human and LLM-judge evaluations differed the most. Importantly, these high-divergence cases did not reflect random disagreement but instead exposed asymmetric interpretive standards between human evaluators and the LLM judge (Table 8).

**Faithfulness discrepancies reflected two opposing but systematic patterns.**

First, in cases where human ratings were high but LLM judge scores were low, the LLM judge penalized interpretations that extended beyond literal wording, even when humans viewed those inferences as pragmatically accurate. Explanation text frequently cited the introduction of information "not explicitly stated," indicating a narrow operationalization of faithfulness anchored in surface correspondence rather than intended meaning. Excerpts such as 693 and 86 exemplify this pattern, where affective stance or causal implication was inferred by the model and endorsed by humans but down-scored by the LLM judge (Table 8). Second, high Faithfulness discrepancies also appeared in the reverse direction, where the LLM judge assigned maximal scores to interpretations that humans evaluated as weak or unsupported. In these cases, outputs closely mirrored the topical content of the original utterance but failed to provide meaningful interpretation. Excerpt 153 illustrates this pattern: the LLM judge rewarded apparent alignment with the source text, while human evaluators saw the response as semantically thin. Altogether, these cases indicate that Faithfulness scoring penalized pragmatic inference and over-rewarded surface alignment.

**Correctness discrepancies primarily reflected over-weighting of structural adequacy.** The LLM judge produced high Correctness scores when responses satisfied task format or topical relevance but offered limited interpretive substance. Explanation text emphasized accurate summarization or compliance with prompt expectations, when human evaluators expected deeper engagement with meaning or intent. Excerpt 710 exemplifies this pattern: the LLM judge evaluated the response as fully correct due to structural adequacy, while humans assigned lower scores because the utterance functioned as a topic shift acknowledgment rather than a substantive point made by the interviewee.

Across both metrics, the highest-divergence excerpts reveal that human evaluators and the LLM judge relied on different criteria for judging interpretation quality. Humans prioritized pragmatic alignment, inferential sensemaking, and interpretive depth, whereas the LLM judge emphasized literal correspondence and formal compliance.

**Table 8. Examples of High-Divergence Excerpts for Faithfulness and Correctness**

| Excerpt ID | Metric | Human composite –LLM judge score | Original excerpt (Truncated) | Model interpretive output (Truncated) | LLM judge's explanation of the excerpt (Truncated) | Interpretative pattern |
|---|---|---|---|---|---|---|
| 693 | Faithfulness | 4.89 – 2.00 | "Right, right, oh my god, that's a lot…" | "The interviewer is astonished by how much…" | "Introduces information not explicitly stated." | Human-valued pragmatic inference penalized |
| 86 | Faithfulness | 4.22 – 2.00 | "Because it gave me a negative, and then some support…" | "The system's unreliability was a source of frustration…" | "Adds interpretation beyond original content." | Implied affect and causal meaning down-scored despite human endorsement |
| 153 | Faithfulness | 2.22 – 5.00 | "Mmm, for math teacher. Yeah, okay…" | "The speaker describes extensive experience as an educator…" | "Faithful to the speaker's statement." | Surface alignment over-rewarded despite weak interpretation |

| 710 | Correctness | 2.44 – 5.00 | "So we're talking about a PD that…" | "The interviewee discusses professional development…" | "Accurately summarizes the speaker's point." | Structural adequacy rewarded over interpretive depth |

*Note.* LLM judge scores were rescaled from a 0–1 to a 1–5 range for comparison with human ratings.

## 5. Discussion and Conclusion

In this study, we examined the extent to which LLM-as-judge evaluations align with human judgments of interpretive quality and identified conditions under which alignment and misalignment occur. Across inference models, LLM-as-judge scores captured broad directional trends in human evaluations but diverged substantially in score magnitude. At the metric level, the LLM judge's Coherence metric provided the strongest aggregated signal for model-level interpretive quality, while Faithfulness and Correctness revealed systematic misalignment at the excerpt level.

These findings have direct implications for qualitative research practitioners who are considering the use of LLM-as-judge methods to support the evaluation of interpretive responses. Rather than positioning automated evaluation as a replacement for human judgment, the results clarify how LLM-as-judge signals can be integrated into qualitative workflows in ways that respect the interpretive nature of qualitative analyses. Particularly, the results point to distinct roles for LLM-as-judge methods at different stages of the analytic process, while preserving human evaluators' responsibility for meaning-centered judgment.

### 5.1 LLM-as-Judge as a Tool for Model Elimination Rather Than Final Selection

One practical implication of the findings is that LLM-as-judge ratings appear especially well suited for elimination-oriented decision-making. In this study, the lowest-performing inference models were consistently identified by both human evaluators and the LLM judge, suggesting that automated evaluation can reliably flag models that perform poorly in interpretive tasks. For practitioners, this indicates that LLM-as-judge scores can be used to narrow the candidate set of inference models before deeper interpretive comparison.

Using LLM-as-judge scores in this eliminative role allows researchers to reserve intensive human evaluation for higher-performing models whose interpretive differences are more nuanced and cannot be confidently resolved through automated scoring alone (Roberts et al., 2024; Yue et al., 2025). In this sense, LLM-as-judge methods function effectively as filtering mechanisms, supporting early-stage model decisions.

### 5.2 Structuring Hybrid Evaluation After Model Elimination

Beyond model elimination, the findings of this study retrospectively clarify how automated and human evaluation could be more strategically differentiated once candidate inference models have been narrowed. The results show that LLM-as-judge scores were effective in resolving coarse model-level distinctions, particularly in consistently identifying underperforming models. At the same time, automated evaluation provided limited resolution among the remaining higher-performing models, where aggregated scores often converged or produced conflicting signals across metrics (e.g., high Coherence but lower Faithfulness). In addition, Harmfulness and Stereotyping metrics were not informative for interpretive evaluation and can be excluded in future workflows.

In these situations, differences among models were primarily observable along human-evaluated criteria only—interpretive accuracy, nuance preservation, and interpretive coherence—where automated evaluation no longer offered sufficient resolution. This pattern suggests that the uniform application of human evaluation across all inference models was unnecessary. Instead, the results motivate a workflow in which model-level automated evaluation is first used to eliminate clearly underperforming options (e.g., eliminate the bottom two models in our study), without loss of interpretive rigor.

Importantly, the protocol used in this study—combining stratified human ratings, model-level automated evaluation, and metric-specific explanation text— demonstrates how human evaluation can be conducted more efficiently and productively when guided by empirical performance patterns observed in the data. Building on the present findings, this protocol can be upgraded to first use model-level automated evaluation to eliminate clearly underperforming models, and then apply stratified human evaluation only to the remaining candidates. In this way, human effort is concentrated where it is most analytically informative. Consistent with prior work framing LLMs as collaborators that support human interpretive judgment (Shin, 2025), this results-based refinement offers a practical evaluation strategy that preserves interpretive quality while making more effective use of limited human evaluation resources.

**5.3 Preserving Human Judgment While Leveraging Automated Signals**

Finally, the findings highlight that human judgment remains indispensable for evaluating interpretive depth, nuance, and pragmatic meaning in qualitative analysis (Chen et al., 2024; Khan et al., 2024). Although LLM-as-judge ratings provide useful signals about structural adequacy and literal alignment, they do not fully reflect the inferential reasoning and contextual sensitivity that human evaluators bring to interpretive work. In this respect, the explanation text accompanying automated scores is especially valuable: it makes the basis of each score transparent, enabling practitioners to examine how automated metrics are being applied and to help assess whether those metrics align with their analytic criteria.

In summary, these results reinforce an evidence-grounded approach to automation in qualitative research. Rather than treating LLM-as-judge methods as substitutes for human evaluation, their most appropriate role is to augment human judgment by offering provisional signals that require contextual interpretation and validation (Liu & Sun, 2025; Roberts et al., 2024). Consistent with evidence that LLM-as-judge performance is highly task-dependent (Jiang et al., 2025), automated evaluations should be calibrated to specific analytic purposes rather than assumed to be universally reliable. By clarifying both the utility and the limits of LLM-as-judge methods, this study provides practical guidance for integrating automated assessment into qualitative research in ways that preserve interpretive integrity while making strategic use of computational support.

**Data and Code Availability.** Because the interview data contain identity-bearing information, the raw excerpts and derived model outputs cannot be publicly released. However, all analysis scripts used to compute alignment statistics, rescale automated scores, and identify high-discrepancy excerpts are publicly available at https://github.com/song9fl/llmAsJudge.